# Fractal Characterization of Low-Correlation Signals in AI-Generated Image Detection

Wenwei Xie[1] (IEEE Member), Jie Yin[2]*, Lu Ma[3], Xuansong Zhang[4], Wenjing Zhang[5]

(1. Trend Micro Incorporated, Nanjing 210000, Jiangsu, China)
(2. Jiangsu Police Institute, Nanjing 210031, Jiangsu, China)
(3. BASF Incorporated, Nanjing 210000, Jiangsu, China)
(4. BMW Incorporated, Nanjing 210012, Jiangsu, China)
(5. Geely Automobile Research Institute, Ningbo 315000, Zhejiang, China)

**Abstract:** AI-generated imagery has reached near-photorealistic fidelity, yet this technology poses significant threats to information security and societal trust. Existing deepfake detection methods often exhibit limited robustness in open-world scenarios. To address this limitation, this paper investigates intrinsic discrepancies between synthetic and authentic images from a signal-level perspective. Our analysis reveals that low-correlation signals serve as distinctive markers for differentiating AI-generated imagery from real photographs. Building on this insight, we introduce a novel method for quantifying these signals based on fractal theory. By analyzing the fractal characteristics of low-correlation signals, our method effectively captures the subtle statistical anomalies inherent to the synthesis process. Extensive experimental results demonstrate the method's robustness and superior detection performance. This work emphasizes the need to shift research focus to a new signal-level direction for Deepfake detection.

## 1. Introduction

Deepfake technology, powered by advanced generative models ranging from Generative Adversarial Networks (GANs) to modern Diffusion Models, enables the creation of highly realistic forged facial content, including face-swapped videos and synthetic talking faces. While these techniques offer creative value for media production and visual effects, their unregulated misuse has in many contexts become a serious threat to information integrity, social trust, and individual privacy. Deepfakes are increasingly used to spread disinformation, commit identity fraud, and fabricate malicious materials. As generative AI continues to evolve, Diffusion-based tools now produce near-indistinguishable synthetic content, and generation improvements have in many cases outpaced the development of reliable detectors - making robust deepfake detection a critical research priority in computer vision and cybersecurity.

Early Deepfake detection methods relied on Convolutional Neural Network (CNN) architectures such as XceptionNet and EfficientNet, which achieved strong performance on closed-domain benchmarks by exploiting local texture and artifact cues. However, pure CNN approaches have inherent limitations: they tend to overfit dataset-specific biases, resulting in significant cross-dataset performance degradation [4], and they struggle to generalize to novel generation techniques encountered in the wild.

Several transformative advances. Vision Transformer (ViT)-based methods emerged as a leading approach between 2024 and 2025, delivering far superior cross-dataset generalization compared to traditional CNNs [4]. Complementary breakthroughs include parameter-efficient methods such as GenD, which achieves strong cross-dataset detection accuracy with only 0.03% of parameters fine-tuned [1], and frequency-domain approaches such as FreqNet realize efficient end-to-end feature learning in the frequency space, yielding lightweight detection models with outstanding cross-dataset generalization performance [2]. The release of large-scale, multi-dimensional datasets such as MultiFF - featuring over 900,000 images generated by 80+ generation algorithms - has also established more realistic evaluation benchmarks, bridging the gap between lab-controlled testing and real-world deployment [3].

Despite substantial advances, critical challenges remain that impede practical deployment. Content synthesized by diffusion models exhibits artifact distributions that often evade detectors trained on earlier generation types; in addition, adversarial attacks, severe class imbalance in operational data, post-processing and compression, and the requirements for real-time, low-latency inference further complicate robust deployment.

## 2. Related Works

In open-world settings, we tested multiple detection models and found that their performance is insufficient, indicating a critical gap between current methods and real-world applicability. The low detection accuracy arises from two main factors. On one hand, the rapid advancement of image generation techniques has rendered synthetic images nearly indistinguishable from authentic ones. On the other hand, the primary object signals within images mask the subtle discriminative features that distinguish real from fake, limiting the effectiveness of current detection models.

This paper categorizes image signals according to their correlation with the main object into high- and low-correlation groups and assesses their discriminative power for distinguishing real from fake images. Based on these insights, we propose a fractal-based approach to quantify low-correlation signals, establishing a novel signal-level criterion for generated image detection.

Figure 1 illustrates the complete workflow. Real and fake images are first cropped to extract face regions, which are then analyzed using Principal Component Analysis (PCA) to separate high- and low-correlation signals. Fractal features are extracted from the resulting residual images, culminating in a final validity assessment.

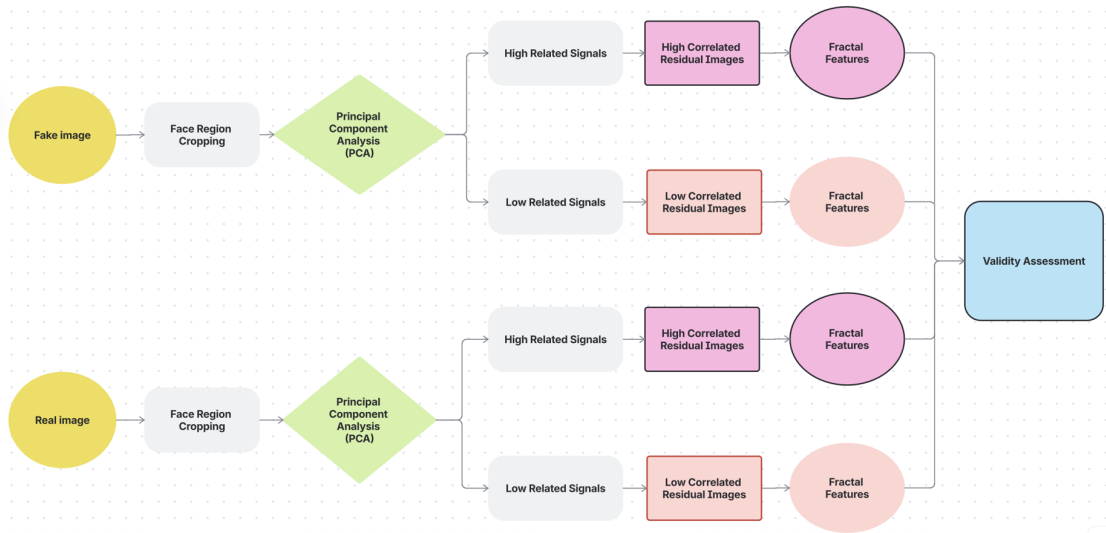

Figure 1: Research Workflow Diagram

## 3. Terminology

To clarify the concepts used in this paper, the following terms are defined:

**Main Object**: The primary object or region in an image that conveys the main semantic information, attracts human visual attention, and exhibits salient features relevant to computer vision tasks.

**Image Semantics**: The conceptual information represented by an image, including the category and attributes of the main object and its relationships with the surrounding environment or other objects.

**High-Correlation Signals**: Signals strongly correlated with the main object and its semantics, contributing significantly to object recognition or semantic understanding.

**Low-Correlation Signals**: Signals weakly correlated with the main object and its semantics, contributing minimally to object recognition or semantic understanding, but potentially capturing subtle differences between generated and real images.

### 4. Contributions

a) Demonstrated the high discriminative power of low-correlation signals for distinguishing AI-generated images from real images.

b) Proposed a fractal-based method to quantify low-correlation signals, establishing a novel signal-level criterion for generated images detection.

c) Validated the effectiveness of the proposed method through extensive experiments.

### 5. Methodology

#### 5.1 Problem Formulation

Current research on generated image detection faces several key challenges. Generated images are produced by diverse sources and multiple generation models, and the main objects within these images often vary. In existing detection methods, dominant signals from the main objects can mask the subtle discrepancies between generated and real images, thereby constraining the generalization of detectors across different models.

#### 5.2 Signal Analysis

Existing generation models, such as GANs and diffusion models, primarily learn signals related to the main object and its semantics during training, while non-main object signals—such as background textures and fine-grained noise patterns—are often ignored. Accordingly, image signals can be categorized into high-correlation signals, which are strongly associated with the main object and its semantics, and low-correlation signals, which exhibit weak correlation. AI-generated images closely resemble real images in high-correlation signals but differ in low-correlation signals, making the latter a reliable cue for AI-generated image detection.

#### 5.3 Principal Component Analysis (PCA)

Principal Component Analysis (PCA) is a dimensionality reduction algorithm that projects data onto orthogonal principal component directions via a linear transformation, with each component capturing the maximum variance.

In this study, PCA is applied to analyze the correlation between image signals and the main object. The top $N$ principal components represent high-correlation signals, while the remaining components correspond to low-correlation signals. The residual image is reconstructed without the top $N$ components, emphasizing low-correlation signals, and providing a means to analyze signal-level differences between generated and real images.

Let the original image point set be $x \in \mathbb{R}^{H \times W}$, After PCA decomposition, the eigenvector matrix $U = [u_1, u_2, ..., u_{HW}]$ is obtained (arranged in descending order of eigenvalues). The reconstructed image retaining the top $N$ principal components is $\hat{x}^{(N)}$, where $u_k^T x$ is the projection coefficient of the original image on the $k-th$ principal component, and $\hat{x}^{(N)}$ is the image vector reconstructed based on the top $N$ principal components.

$$\hat{x}^{(N)} = \sum_{k=1}^{N} \left( u_k^T x \right) u_k \qquad (1)$$

The residual image is $R^{(N)}$, representing the residual between the original image and the reconstructed image retaining the top $N$ principal components.

$$R^{(N)} = x - \hat{x}^{(N)} \qquad (2)$$

#### 5.4 Fractal Theory

Fractal theory is employed to characterize complex structures and self-similar patterns in natural or artificial images. The fractal dimension (FD) measures the geometric complexity and texture roughness of images, providing a quantitative description of microscopic structural features. The multifractal spectrum captures local structural variations at different scales, revealing heterogeneity in complex texture patterns. Differences between images are quantified by computing the fractal dimension, multifractal spectrum, and related statistics such as lacunarity and information entropy.

**Fractal Dimension:** The fractal dimension $D$ can be calculated by the box-counting method, where $N(\epsilon)$ represents the number of boxes with side length $\epsilon$ covering the image.

$$D = \lim_{\epsilon \to 0} \frac{\log N(\epsilon)}{\log(1/\epsilon)} \quad (3)$$

**Lacunarity:** Lacunarity is an indicator in fractal analysis used to quantify the non-uniformity of the spatial distribution of texture gaps in images and can reflect the roughness and heterogeneity of structures in images.

Let the image or binary image $I$ be divided into boxes of size $\epsilon \times \epsilon$, where the sum of pixel values in the $i-th$ box is $s_i$, and the total number of boxes is $N_\epsilon$. Lacunarity $\Lambda(\epsilon)$: the larger $\Lambda(\epsilon)$, the more non-uniform the image and the more gaps.

$$\Lambda(\epsilon) = \frac{\text{Var}(s_i)}{[\mathbb{E}(s_i)]^2} + 1 \quad (4)$$

Where:

$$\text{Var}(s_i) = \frac{1}{N_\epsilon} \sum_{i=1}^{N_\epsilon} (s_i - \mathbb{E}(s_i))^2 \quad (5)$$

$$\mathbb{E}(s_i) = \frac{1}{N_\epsilon} \sum_{i=1}^{N_\epsilon} s_i \quad (6)$$

**Shannon Entropy：** Information entropy is used to quantify the uncertainty or complexity of the grayscale distribution of images. Let the set of pixel values of a grayscale image be $X = x_1, x_2, \dots, x_n$, with pixel values in the range $[0, L-1]$, and probability distribution $p(x_i)$. The image information entropy $H(X)$ is:

$$H(X) = -\sum_{i=0}^{L-1} p(x_i) \log_2 p(x_i) \quad (7)$$

Where:

$$p(x_i) = \frac{\text{Number of pixels with value } x_i}{\text{Total number of pixels}} \quad (8)$$

**Multifractal Spectrum (MFS):** A single Fractal Dimension (FD) can only describe the overall complexity of an image, while the multifractal spectrum can reveal the complexity of images in different local regions.

Let the measure of the image be $\mu$. Divide the image into boxes, with each box size $\epsilon$ and the $i-th$ box measure $\mu_i(\epsilon)$. When $\epsilon \to 0$, the relationship between the $q-\text{order}$ fractal moment (Partition Function) and the scale $\epsilon$ satisfies a power law, where $\tau(q)$ is the multifractal index function.

$$\chi(q,\epsilon) = \sum_i \mu_i(\epsilon)^q \sim \varepsilon^{\tau(q)} \quad (9)$$

$\alpha_q$ represents local intensity, and $f(\alpha)$ represents the corresponding dimension.

$$\alpha_q = \lim_{\epsilon \to 0} \frac{\log \mu_i(\epsilon)}{\log \epsilon} \quad (10)$$

$$\alpha(q) = \frac{d\tau(q)}{dq} \quad (11)$$

$$f(\alpha) = q\alpha - \tau(q) \quad (12)$$

### 5.5 Experimental Setup

This paper conducts experiments with the dataset 1-million-fake-faces for fake images (https://www.kaggle.com/datasets/tunguz/1-million-fake-faces), dataset FFHQ for real images (https://www.kaggle.com/datasets/arnaud58/flickrfaceshq-dataset-ffhq). PCA is applied to extract low-correlation signals, and statistics such as information entropy, fractal dimension (FD), multifractal spectrum (MFS) and Lacunarity are employed to characterize the images.

This study analyzes differences in classification performance between real and fake images before and after suppressing main-object signals. The source code for these experiments is available at https://github.com/jim-xie-cn/Research-Deepfake.

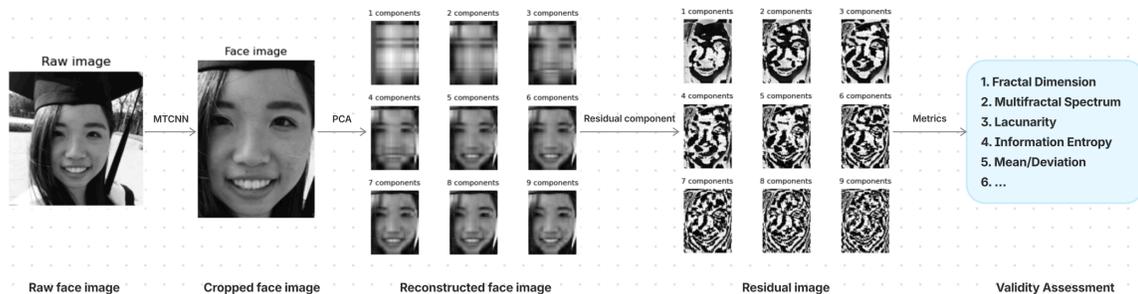
Figure 2: Experimental Processing Workflow.

As shown in Figure 2, the experimental workflow starts with raw face images. After cropping to extract the face regions, PCA is applied to obtain the principal components, and the images are reconstructed using the top $N$ components. Residual images are then computed, and various metrics are extracted, including fractal dimension, multifractal spectrum, lacunarity, information entropy, and statistics (mean and standard deviation). The workflow consists of the following four steps:

**Dataset Construction**: 10,000 real images and 10,000 generated images are selected from the 1-million-fake-faces dataset. Face regions are extracted using MTCNN to construct the experimental dataset.

**Residual Image Generation:** PCA decomposition is performed on the face images. The image is reconstructed based on the top $N$ principal components, and the residual image is obtained by computing the difference between the original image and the reconstruction.

**Statistical Feature Calculation**: Fractal dimension, multifractal spectrum, lacunarity, information entropy, and statistics (mean and standard deviation) are extracted from the residual images.

**Validity Assessment:** Differences in the extracted statistical features between real and generated residual images are compared to evaluate their discriminative power. Kolmogorov–Smirnov (KS) tests were conducted to examine the statistical properties of the extracted features—mean, standard deviation, fractal dimension, information entropy, MFS and lacunarity—for real and fake images. First, the KS test was used to assess whether the distributions of these features conformed to a normal distribution. Subsequently, the KS test was applied to compare the feature distributions between real and fake images.

## 6. Results

From the cropped face images, the top 1 to top 32 principal components were extracted, and the corresponding residual images were computed. Statistical features were then calculated for these residual images and compared with those of the original cropped face images. The resulting observations are summarized as follows:

**Raw Image Assessment (before PCA):**

The statistical features of raw images, prior to PCA processing, were analyzed to evaluate their distributions and discriminative potential. Figure 3 illustrates the distributions of fractal dimension (FD), information entropy, mean, and standard deviation for real and fake images. Figure 4 illustrates the mean multifractal spectrum (MFS), including scaling exponents ($\tau$), singularity indices ($\alpha$), generalized dimensions ($d$), and multifractal spectrum ($f(a)$). Figure 5 illustrates the mean lacunarity at different scales. Across all these features, There is no clear discriminative patterns are observed between real and fake images, suggesting that the statistical properties of raw images alone are insufficient to differentiate generated images from authentic ones.

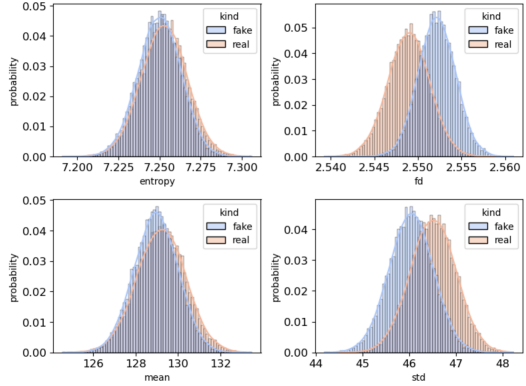

Figure 3: Statistics Distribution of Raw Images.

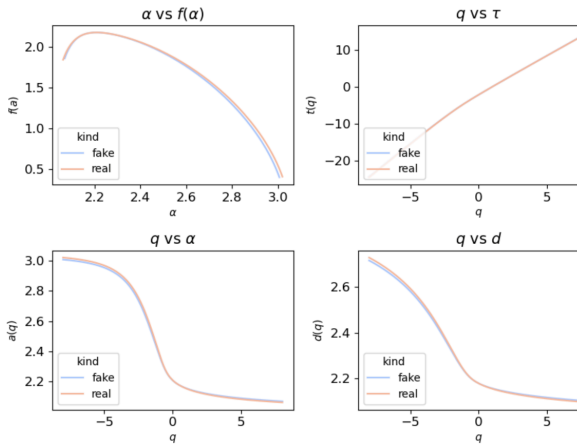

Figure 4: Mean MFS of Raw Images.

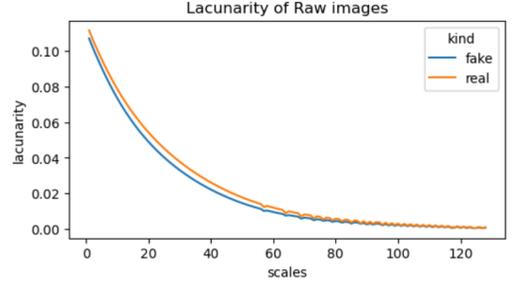

Figure 5: Mean Lacunarity of Raw Images.

As summarized in Table 1, the KS statistics for real images range from 0.00437 to 0.00545, with corresponding p-values between 0.59 and 0.84. For fake images, KS statistics range from 0.0037 to 0.0046, with p-values between 0.80 and 0.95. These results indicate that all features approximately follow normal distributions.

When comparing the distributions of real versus fake images, the KS statistics reveal significant differences across all features (FD: 0.537, entropy: 0.089, mean: 0.104, std: 0.380), with p-values approximately 0.0. This apparent contradiction with the visual similarity shown in Figure 3 can be attributed to the dominance of high-correlation signals, which largely mask subtle differences in the raw images. Despite each feature individually exhibiting approximate normality, no clear discriminative patterns are apparent, suggesting that subtle differences between real and generated images are obscured by the main-object signals.

Table 1: KS Test for Real and Fake Image of Raw Images

| Feature | Real vs Normality | | Fake vs Normality | | Real vs Fake | |
|---|---|---|---|---|---|---|
| | D | p_value | D | p_value | D | p_value |
| FD | 0.004371 | 0.837706 | 0.003700 | 0.946161 | 0.537356 | <0.001 |
| Entropy | 0.005248 | 0.638335 | 0.004563 | 0.797457 | 0.089238 | <0.001 |
| Mean | 0.004933 | 0.713254 | 0.004137 | 0.004037 | 0.104401 | <0.001 |
| Std | 0.005448 | 0.590825 | 0.881943 | 0.89902 | 0.379718 | <0.001 |

**Residual Image Assessment (after PCA):**

Residual images were generated by reconstructing face images without the top $N$ principal components, where $N$ ranged from 24 to 32. Figure 6 illustrates the distributions of fractal dimension (FD), information entropy, mean, and standard deviation for real and fake images based on these residual images

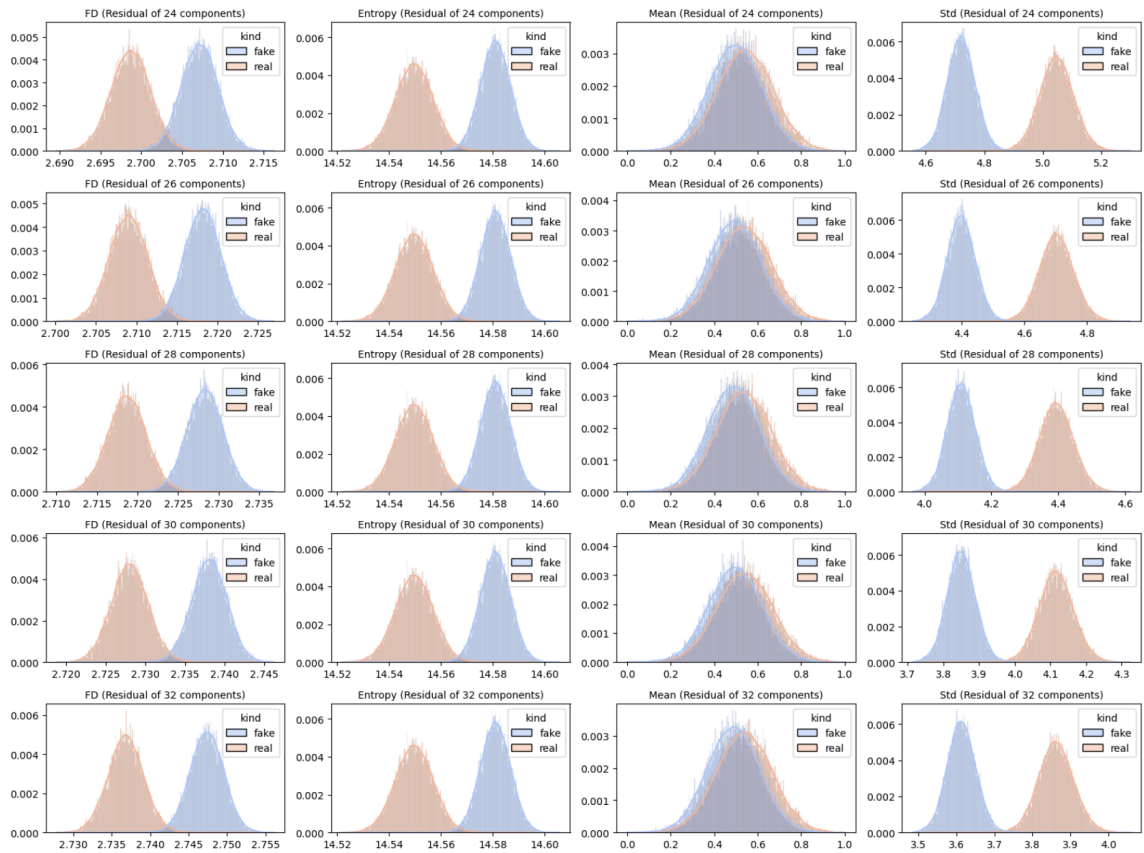

Figure 6: Statistics Distribution of Residual Images.

Figure 7 illustrates the mean multifractal spectrum (MFS) of residual images, with the horizontal axis representing singularity indices ($\alpha$) and the vertical axis representing multifractal spectrum $f(a)$. It can be observed that as more top principal components (high-correlation signals) are removed, the discriminative patterns between real and fake images become increasingly pronounced.

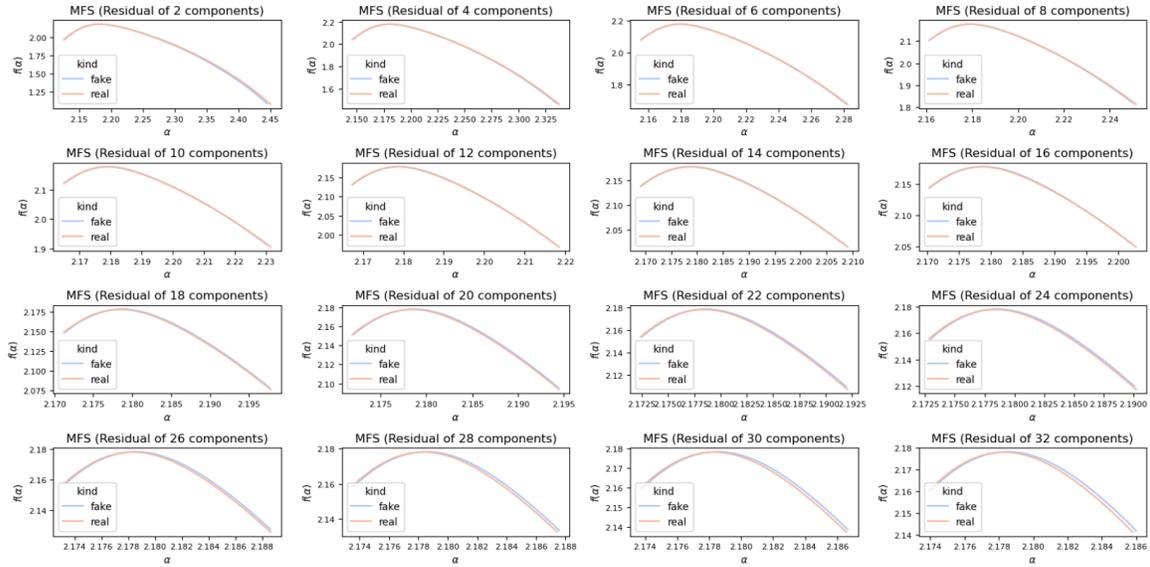

Figure 7: Mean MFS of Residual Images.

Figure 8 illustrates the heatmap of KS statistics for MFS-based features across different scales ($q$) and numbers of removed PCA components. This visualization highlights clear patterns: the distinction between real and fake images strengthens as more high-correlation components are suppressed, confirming the enhanced discriminative power of low-correlation signals.

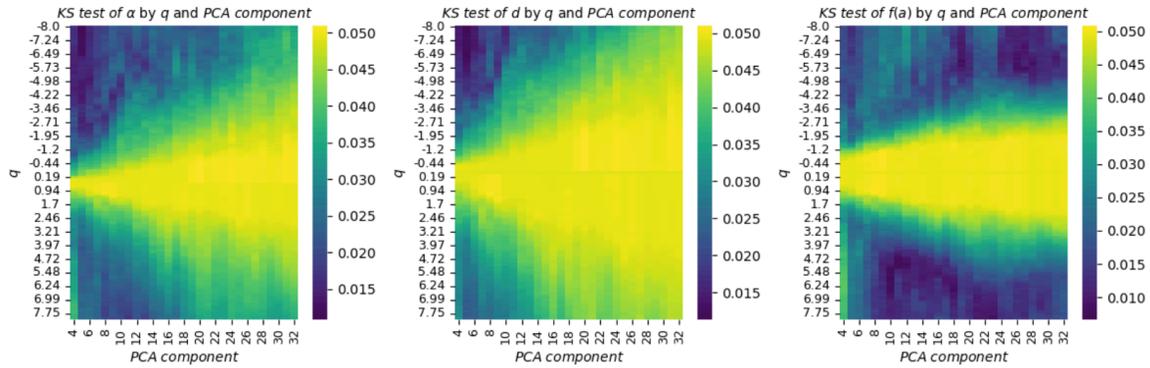

Figure 8: KS test for MFS Statistics of Residual Images.

Figure 9 illustrates the mean lacunarity of residual images at different scales. Unlike FD, entropy, and MFS, lacunarity does not exhibit significant discriminative ability, indicating that it may not effectively capture subtle differences between real and generated images in the residual domain.

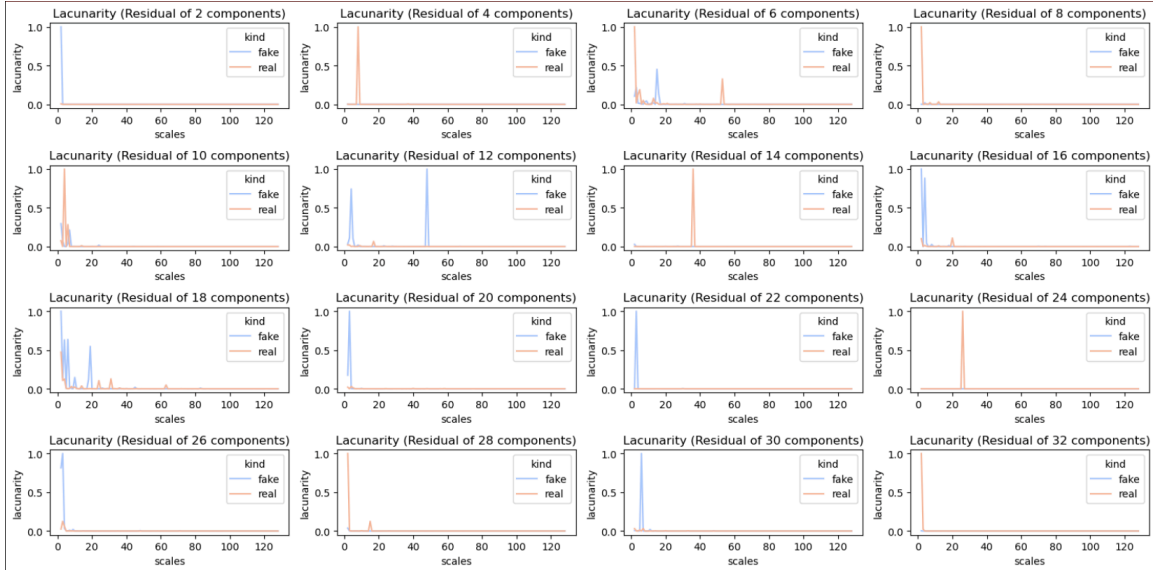

Figure 9: Mean Lacunarity of Residual Images.

Kolmogorov–Smirnov (KS) tests were conducted to quantitatively evaluate the differences between real and fake residual images (Table 3). The KS statistics for FD and entropy consistently exceed 0.93 across all residual images, with corresponding p-values ≈ 0.00, indicating significant differences between real and fake images. Standard deviation also shows notable differences, with KS statistics close to 1.000 and p-values ≈ 0.00. (The statistic mean shows limited discriminative power between real and fake residual images, likely due to the symmetric, approximately normal distributions produced by generative models). These results demonstrate that suppressing high-correlation signals and extracting low-correlation signals markedly enhances the discriminative power of the extracted features (which are consistent with those shown in Figure 5).

Table 3: KS Test for Real and Fake Image of Residual Images

| Residual | FD | | Entropy | | Mean | | Std | |
|---|---|---|---|---|---|---|---|---|
| | D | p_value | D | p_value | D | p_value | D | p_value |
| 24 components | 0.93613 | <0.001 | 0.98791 | <0.001 | 0.15010 | <0.001 | 1.0000 | <0.001 |
| 26 components | 0.95778 | <0.001 | 0.98780 | <0.001 | 0.13847 | <0.001 | 1.0000 | <0.001 |
| 28 components | 0.96707 | <0.001 | 0.98790 | <0.001 | 0.15303 | <0.001 | 1.0000 | <0.001 |
| 30 components | 0.97699 | <0.001 | 0.98680 | <0.001 | 0.13807 | <0.001 | 1.0000 | <0.001 |
| 32 components | 0.98501 | <0.001 | 0.98791 | <0.001 | 0.16164 | <0.001 | 0.9998 | <0.001 |

## 7. Conclusion

In this study, we found the role of low-correlated signals in images, revealing their significance in identifying real and AI-generated images, and thereby providing a new research direction for deep fake detection.

Building on this insight, this paper proposes a fractal feature analysis method based on low-correlated signals. By analyzing the statistical features of residual images, this method effectively captures differences between real and AI generated images, providing a novel signal-level basis for AI-generated image detection.

Theoretically, the proposed approach is not limited to face images identification but can be applied to all AI-generated image detection tasks. In practice, the extracted low-correlated signals can be used to generate fingerprints of generative models for identifying AI-generated images.

Building on this study, a PCA and fractal-based workflow for real/fake image recognition can be outlined as follows:

1. **Residual Image Construction**: PCA is employed to extract low-correlation signals and generate residual images.

2. **Feature Extraction:** Statistical features of residual images, such as information entropy and fractal dimension are extracted for discrimination between real and AI-generated images.

3. **Detailed Analysis with MFS**: The multifractal spectrum is employed to extract more detailed structural features, enabling finer-grained recognition between real and generated images.

This workflow demonstrates that combining PCA with fractal theory provides a robust and generalizable framework for distinguishing real and AI-generated images, highlighting the potential of low-correlation signals as a powerful discriminative tool.

## Authors

| | |
|---|---|
| 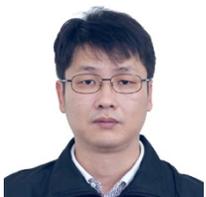 | **Wenwei Xie**: IEEE Member, received a master's degree in MBA from the Nanjing University of Aeronautics and Astronautics (NUAA), Nanjing, China in 2013. He is currently working toward network security in Trend Micro Incorporated, Nanjing, China. His research interests include Cyber security, Artificial Intelligence and Computer vision. His e-mail is jim.xie.cn@outlook.com. |
| 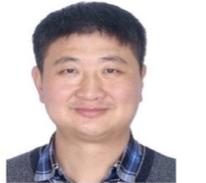 | **Jie Yin**: received M.S. degree in Software Engineering from Nan- Jing University of Science and Technology, Nanjing, China in 2008. He is currently a Senior Engineering with the Department of Computer Information and Cyber Security, Jiangsu Police Institute, Nanjing, China. His recent research interests include machine learning, big data, and network security. His e-mail is yinjiejspi@163.com. |
| 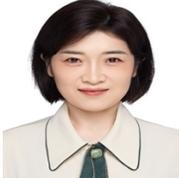 | **Lu Ma**: received the master's degree in electronic engineering from Southeast University, Nanjing, China in 2013. She is a Senior Digitalization Expert with the Digital Hub China at BASF in Nanjing, China. Her research interests include application security, network security, AI technology and computer vision. Her e-mail is hml163mail@163.com. |
| 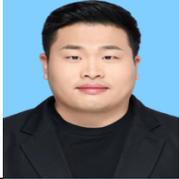 | **Xuansong Zhang**: received the master's degree of Statistics from George Washington University, US in 2017. He is currently working toward the Data engineer in BMW Cooperation, Nanjing, China. His research interest includes Time Series, AI Technology, and computer vision. His e-mail is zhyzxsw@126.com. |
| 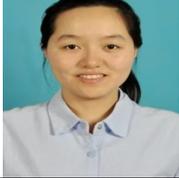 | **Wenjing Zhang**: received the B.S. degree from Anhui University, Hefei, China. She is currently engaged in the field of Internet of Vehicles (IoV) Big Data. Her research interests include IoV big data analysis, machine learning, artificial intelligence, and model security. Her e-mail is clairedyx@126.com. |